\documentclass[10pt,letterpaper]{article}

\usepackage[utf8]{inputenc}
\usepackage[T1]{fontenc}
\usepackage[margin=1in]{geometry}

\usepackage{newtxtext, newtxmath}
\usepackage{microtype}
\usepackage{enumitem}

\usepackage{graphicx}
\usepackage{amsmath}
\usepackage{booktabs} 
\usepackage{algorithm}
\usepackage{algorithmicx}
\usepackage{algpseudocode}
\usepackage{multirow}

\usepackage{authblk}

\usepackage{natbib}
\usepackage[colorlinks,allcolors=blue,breaklinks]{hyperref} 

\title{\bfseries Learning to Seek Evidence: A Verifiable Reasoning Agent with Causal Faithfulness Analysis}

\author[1]{Yuhang Huang}
\author[2]{Zekai Lin}
\author[3,*]{Fan Zhong}
\author[3,4,5,*]{Lei Liu}

\affil[1]{Institute of Biomedical Science, Fudan University, Shanghai, China}
\affil[2]{Fudan University, Shanghai, China}
\affil[3]{Intelligent Medicine Institute, Fudan University, Shanghai, China}
\affil[4]{Shanghai Institute of Infectious Disease and Biosecurity, Fudan University, Shanghai, China}
\affil[5]{Shanghai Institute of Stem Cell Research and Clinical Translation, Fudan University, Shanghai, China}
\affil[*]{\textit{Corresponding authors}: zonefan@163.com, liulei@fudan.edu.cn}

\date{} 

\begin{document}

\maketitle

\begin{abstract}
Explanations for AI models in high-stakes domains like medicine often lack verifiability, which can hinder trust. To address this, we propose an interactive agent that produces explanations through an auditable sequence of actions. The agent learns a policy to strategically seek external visual evidence to support its diagnostic reasoning. This policy is optimized using reinforcement learning, resulting in a model that is both efficient and generalizable. Our experiments show that this action-based reasoning process significantly improves calibrated accuracy, reducing the Brier score by 18\% compared to a non-interactive baseline. To validate the faithfulness of the agent's explanations, we introduce a causal intervention method. By masking the visual evidence the agent chooses to use, we observe a measurable degradation in its performance ($\Delta$Brier=+0.029), confirming that the evidence is integral to its decision-making process. Our work provides a practical framework for building AI systems with verifiable and faithful reasoning capabilities.
\end{abstract}

\section{Introduction}
\label{sec:intro}

Deep learning models have achieved remarkable success in medical image analysis, yet their "black box" nature remains a major barrier to clinical adoption~\cite{Chen2022HCDXAI, JMIRAI2024HumanCenteredXAIReview}. To build trust, a common practice is to generate post-hoc saliency maps, such as Grad-CAM~\cite{Selvaraju2017GradCAM, Zeiler2014Visualizing}, to highlight regions deemed important by the model. However, the reliability of these heatmaps has been repeatedly questioned. Studies show they can be insensitive to model parameters or data labels~\cite{Adebayo2018Sanity}, exhibit "Clever Hans" behaviors by focusing on shortcuts~\cite{Lapuschkin2019CleverHans}, and often demonstrate poor localization accuracy in rigorous radiological evaluations~\cite{Zhang2024RAI_SaliencyTrust, Yanagawa2023RAI_Commentary}. This makes saliency maps alone insufficient for accountable, high-stakes decision support.

In response to these limitations, we argue for a paradigm shift: from post-hoc rationalization to \textbf{verifiable reasoning-in-action}. We operationalize this paradigm by architecting an interactive agent around a \textbf{Vision-Language Model (VLM)}, inspired by the recent agentic reasoning trend~\cite{Yao2023ReAct, Schick2023Toolformer}. We leverage the VLM's native ability to jointly process images and text to generate an explicit, step-by-step reasoning trace. To structure this process, we model the diagnostic workflow as a transparent loop between a \textbf{Hypothesis Box (H-Box)}, where the agent maintains and updates its beliefs, and a \textbf{Probe \& Ground (P\&G)} action for evidence validation. This transforms the opaque process of diagnosis into a transparent, transactional trace.

The cornerstone of our framework is the \textbf{P\&G} action, which creates a tight feedback loop between hypothesis and evidence. When the agent decides to probe the image, it invokes a dedicated external tool---which we term the \textbf{Knowledge-Based Confidence Scorer (KBCS)}---that analyzes the image to produce a candidate Region of Interest (ROI) and returns a calibrated numerical confidence score. The agent then integrates this feedback into its \textbf{H-Box}, turning explanation into a dynamic interaction rather than a static caption. Crucially, our entire system is designed for accessibility. By leveraging 4-bit quantization and lightweight reinforcement learning~\cite{Hu2021LoRA, Dettmers2023QLoRA}, it is trainable end-to-end on a single 24GB GPU.

\begin{figure*}[t]
    \centering
    \includegraphics[width=0.7\textwidth]{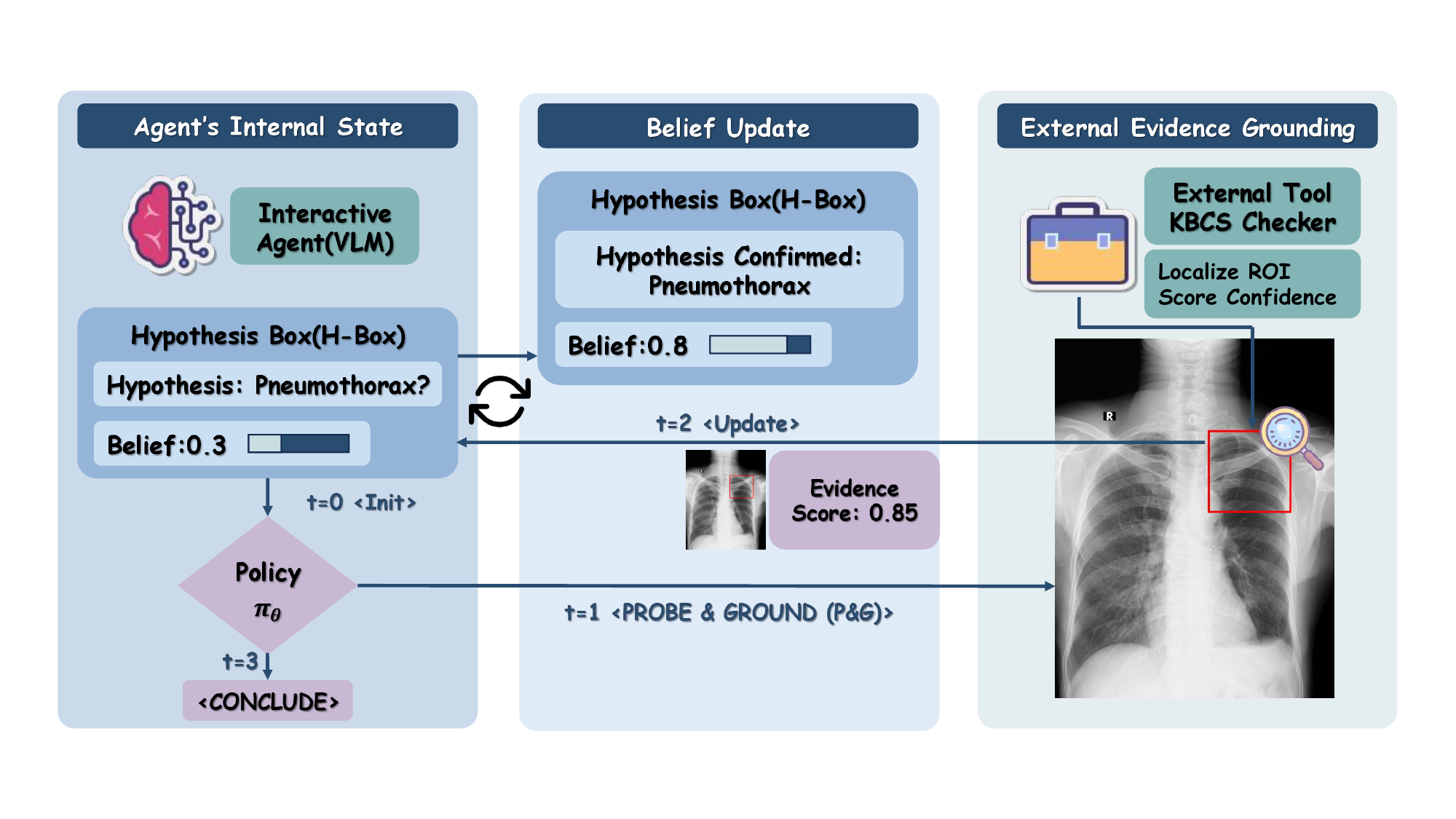} 
    \caption{Overview of our verifiable reasoning framework. The agent iteratively refines its belief within a \textbf{Hypothesis Box (H-Box)} by executing a \textbf{--P\&G)} action. This action invokes an external tool (KBCS) to ground the hypothesis in visual evidence (an ROI and a score), creating an auditable, step-by-step diagnostic trace.}
    \label{fig:overview}
\end{figure*}

To validate our framework, we move beyond passive metrics and champion a protocol for interventional evaluation. We causally probe the faithfulness of explanations using occlusion tests~\cite{Fong2017MeaningfulPerturbations, Petsiuk2018RISE, Hooker2019ROAR}, measuring the model’s output change when its claimed evidence is masked. This, combined with metrics for calibration and consistency, forms a comprehensive suite for assessing whether an explanation is truly grounded in visual evidence.

Our main contributions are:
\begin{itemize}
    \item A novel reasoning framework that models diagnosis as a transparent loop between a \textbf{H-Box} for belief updates and a \textbf{P\&G} action for evidence validation.
    \item A verifiable evidence-grounding mechanism where the \textbf{P\&G} action invokes an external tool to return a calibrated confidence score, directly informing the agent's policy.
    \item A low-compute RL alignment strategy that enables training of a 4-bit quantized agent on a single 24GB GPU, making verifiable reasoning accessible on commodity hardware.
    \item A comprehensive interventional evaluation protocol that unifies occlusion-based faithfulness tests with calibration metrics to rigorously validate that explanations are causally linked to decisions.
\end{itemize}
\section{Related Work}
\label{sec:related}

\paragraph{Intrinsic vs. Post-hoc Explanations.}
The debate on model explainability often centers on two paradigms. The dominant approach is post-hoc saliency maps~\cite{Selvaraju2017GradCAM, Zeiler2014Visualizing}, which rationalize a black-box model's decision. However, their faithfulness is heavily contested, with studies revealing sensitivity issues~\cite{Adebayo2018Sanity}, reliance on spurious correlations~\cite{Lapuschkin2019CleverHans}, and poor clinical localization~\cite{Zhang2024RAI_SaliencyTrust, Yanagawa2023RAI_Commentary}. The alternative is inherently interpretable models, such as those using concept bottlenecks~\cite{Koh2020ConceptBottleneck, Chen2020ConceptWhitening}, which constrain the model's internal structure. Our work proposes a third path: \textbf{verifiable process-based explanations}. Instead of interpreting a model's internal state, we make its external reasoning process—the sequence of actions—the primary object of audit. This shifts the focus from "what the model saw" to "how the model decided."

\paragraph{Agentic Frameworks for Visual Reasoning.}
Recent advances have endowed VLMs with agentic capabilities, enabling them to interleave reasoning with actions to solve complex tasks~\cite{Yao2023ReAct, Schick2023Toolformer}. In the visual domain, systems like ViperGPT~\cite{Suris2023ViperGPT} and Visual ChatGPT~\cite{Wu2023VisualChatGPT} orchestrate a suite of vision tools to fulfill user requests. While these systems focus on maximizing task-completion performance, our primary goal is different: we constrain the agent's behavior to generate a faithful and auditable diagnostic trace. We design a compact, domain-specific action space not for general-purpose ability, but for verifiable clinical reasoning. Our agent's actions are not just steps towards an answer; they \emph{are} the explanation.

\paragraph{Aligning Reasoning Processes with Reinforcement Learning.}
RL has become a powerful paradigm for aligning large language models with desired behaviors, such as helpfulness and harmlessness, famously demonstrated by RLHF~\cite{Ouyang2022InstructGPT}. This principle of alignment can be extended beyond conversational preference to specific reasoning traits. Our work applies this concept to instill a policy of \textbf{verifiable evidence-seeking}. We use RL not to maximize a downstream task score directly, but to reward the agent for taking actions that ground its beliefs in evidence before committing to a conclusion. While concurrent work like SEAL~\cite{Zweiger2025SEAL} uses RL for long-horizon self-improvement, our approach uses a dense, step-wise reward to align the agent's immediate evidence-seeking policy with the goal of verifiable reasoning. This alignment is made computationally feasible through parameter-efficient methods (PEFT) like LoRA~\cite{Hu2021LoRA, Dettmers2023QLoRA}.

\section{Method}
\label{sec:method}

Our framework models the diagnostic process as an iterative reasoning loop performed by a VLM agent. The core idea is to make the agent's decision-making process transparent and verifiable. This is achieved through a cycle of hypothesizing and evidence-seeking, where the agent's actions and their resulting belief updates form an auditable trace. This section details the three pillars of our framework: the agent's reasoning loop (§\ref{sec:method:loop}), the evidence-grounding mechanism (§\ref{sec:method:pg}), and the policy alignment strategy (§\ref{sec:method:rl}).


\subsection{The Agent's Reasoning Loop}
\label{sec:method:loop}
The agent's diagnostic process is an iterative loop, summarized in Algorithm~\ref{alg:reasoning_loop}. At each step, the agent consults its internal state—the \textbf{H-Box}—and makes a fundamental choice: either continue exploring by seeking evidence, or terminate by committing to a decision.

\paragraph{H-Box.}
The H-Box is the container for the agent's internal state, $s_t$. It dynamically tracks the agent's current diagnostic belief, represented as a probability $p_t$, and the history of all prior actions and observations. The belief is initialized with a prior $p_0$, derived either from dataset statistics or an initial VLM query. At each step, the agent observes its H-Box to decide on an action, and the chosen action in turn updates the H-Box. The sequence of states and actions forms the reasoning trace $\tau$.
To enforce verifiable, evidence-based reasoning, any reasoning trace that concludes without at least one evidence-seeking action is considered invalid; its final belief defaults back to the initial prior $p_0$.

\paragraph{Action Space.}
The agent's policy $\pi_\theta(a_t|s_t)$ selects an action $a_t$ from a compact, discrete set. These actions dictate how the agent interacts with its environment and updates its H-Box:
\begin{itemize}
    \item \textbf{\textsc{Probe \& Ground (P\&G)}}: The core evidence-seeking action. It invokes an external tool to find and score visual evidence for the concept $c$. This is a non-terminal action that leads to a belief update.
    \item \textbf{\textsc{Claim}}: A terminal action where the agent asserts its hypothesis with high confidence. This action signals a strong conviction, sharpens the current belief $p_t$ to reflect high confidence, and then concludes the episode.
    \item \textbf{\textsc{Abstain}}: A terminal action indicating uncertainty. The agent concludes the process by explicitly signaling its inability to make a confident decision, setting the final belief $p_{\text{final}}$ to 0.5.
    \item \textbf{\textsc{Stop}}: A general terminal action that concludes the reasoning process based on the H-Box's current belief state, without further modification.
\end{itemize}

\paragraph{Implementation Details.}
The policy is implemented by a 4-bit quantized Qwen2.5-VL-3B. The agent generates actions in a structured JSON format. At decision time, we compute logits for the four action tokens, mask invalid choices (e.g., preventing a \textsc{Claim} before any \textsc{P\&G} action), and form a categorical distribution. A safe-sampling mechanism handles numerical instabilities, defaulting to a uniform distribution or the \textsc{Stop} action if necessary. During training, gradient updates are restricted to LoRA modules and the action-specific token embeddings.

\begin{algorithm}[t]
\caption{Verifiable Reasoning Loop}
\label{alg:reasoning_loop}
\begin{algorithmic}[1]
\State \textbf{Input:} Image $x$, clinical concept $c$, policy $\pi_\theta$
\State \textbf{Output:} Final belief $p_{\text{final}}$, reasoning trace $\tau$

\State Initialize H-Box with prior $p_0$; set $p \leftarrow p_0$, $\tau \leftarrow []$, $\text{probed} \leftarrow \text{false}$

\For{$t = 1, \dots, T_{\text{max}}$}
    \State $a_t \leftarrow \pi_\theta(\text{H-Box}_t)$
    \State Append $(s_{t-1}, a_t)$ to trace $\tau$

    \If{$a_t$ is \textsc{P\&G}}
        \State $\text{ROI}, p_{\text{evidence}} \leftarrow \text{KBCS}(x, c)$
        \State $p \leftarrow \text{FuseEvidence}(p, p_{\text{evidence}})$
        \State $\text{probed} \leftarrow \text{true}$
    \Else
        \If{$a_t$ is \textsc{Claim}} $p \leftarrow \text{SharpenBelief}(p)$ \EndIf
        \If{$a_t$ is \textsc{Abstain}} $p \leftarrow 0.5$ \EndIf
        \State \textbf{break}
    \EndIf
\EndFor

\State // Enforce evidence-based reasoning: no probe means no update.
\State $p_{\text{final}} \leftarrow p$ \textbf{if} $\text{probed}$ \textbf{else} $p_0$
\State \textbf{return} $p_{\text{final}}, \tau$
\end{algorithmic}
\end{algorithm}

\subsection{The P\&G Action: Grounding via the KBCS Tool}
\label{sec:method:pg}

The \textsc{P\&G} action is the cornerstone of our framework, creating a verifiable link between the agent's hypothesis and visual data. Instead of relying on the VLM's opaque internal vision capabilities, this action delegates the evidence-seeking task to a dedicated, independent external tool: the \textbf{KBCS}. The KBCS analyzes the image and returns a tuple $(\text{ROI}, p_{\text{evidence}})$, containing localized visual evidence, or, in some cases, a global evidence score $p_{\text{evidence}}$ without a specific ROI.

\paragraph{Independent Evidence Extraction.}
The KBCS operates on a modular, independent vision stack with a tiered backend system designed to balance efficiency and interpretability.
\begin{itemize}[leftmargin=*, topsep=0pt, itemsep=-2pt]
    \item \textbf{Primary Backend (Global Score):} The tool first attempts to use a highly efficient, fine-tuned vision head built upon a frozen BiomedCLIP~\cite{BiomedCLIP} encoder. This head directly outputs a calibrated global probability $p_{\text{evidence}}$. If this backend provides a score, the KBCS returns it immediately without generating a heatmap or ROI.
    \item \textbf{Fallback Backend (Localized Evidence):} If the primary backend is unavailable or cannot handle the concept, the KBCS falls back to a saliency-based approach. In our experiments, this is a zero-shot Grad-CAM~\cite{Selvaraju2017GradCAM} on a pretrained CXR classifier (e.g., TorchXRayVision). This backend generates a heatmap, from which an ROI is proposed and mapped to the original pixel space. The peak heatmap intensity is then used as the basis for $p_{\text{evidence}}$.
\end{itemize}
To ensure full auditability, the KBCS logs its operational parameters (backend name, model hash, scaling factors) into a \textit{provenance} record for every call.

\paragraph{Score Calibration and Belief Fusion.}
Raw outputs from vision backends require careful calibration. The KBCS ensures any internal score (e.g., a raw logit or a heatmap peak intensity) is processed through a concept-specific calibration layer. This typically involves a learned temperature and bias in the log-odds space:
\begin{equation}
p_{\text{evidence}} = \sigma\left(\frac{m_{\text{raw}}}{T_c} + b_c\right)
\label{eq:calibration}
\end{equation}
where $m_{\text{raw}}$ is the uncalibrated log-odds score and $(T_c, b_c)$ are the learned parameters. This calibrated score is then returned to the agent. The \texttt{FuseEvidence} function (Alg.~\ref{alg:reasoning_loop}, line 9) integrates this new information into the H-Box's belief via a simple weighted average: $p_{t+1} \leftarrow (1-\alpha)p_t + \alpha p_{\text{evidence}}$, where $\alpha$ is a hyperparameter. This closed loop ensures that belief updates are directly and verifiably tied to external, calibrated evidence.

\subsection{Policy Alignment via Reinforcement Learning}
\label{sec:method:rl}

We align the agent's policy $\pi_\theta$ to favor faithful and effective reasoning sequences using a lightweight, conservative policy-gradient procedure, detailed in Algorithm~\ref{alg:rl_alignment}. The goal is to teach the agent \emph{when} to use the costly but informative \textsc{P\&G} action versus when to confidently \textsc{Claim} or \textsc{Stop}.

\paragraph{State Dynamics and Training Proxy.}
The agent's belief $p$ is updated by deterministic transition rules, as described in §\ref{sec:method:pg}. For the \textsc{Claim} action, the \texttt{SharpenBelief} function is $p_{t+1} \leftarrow \sigma(\gamma \cdot \text{logit}(p_t))$ with $\gamma>1$. Crucially, to ensure efficient training, the expensive KBCS tool is only used during evaluation. For the inner loop of RL training, the call to the KBCS is replaced by a lightweight proxy: a direct query to the VLM agent itself to generate a score.

\paragraph{Terminal Reward and Self-Critical Baseline.}
We employ a simple yet effective terminal reward. After an episode concludes with a final belief $p_{\text{final}}$, we compute the negative Brier score as the reward: $R = -(p_{\text{final}} - g)^2$, where $g$ is the ground-truth label. To stabilize training, we use a self-critical baseline. For each training example, we sample $K$ trajectories under the current policy $\pi_\theta$. The baseline is the average reward of these trajectories, $\bar{R} = \frac{1}{K}\sum_k R_k$. The advantage for each trajectory is then $A_k = R_k - \bar{R}$, rewarding actions that lead to above-average outcomes. All advantages from a minibatch are then standardized.

\paragraph{Conservative Policy Update (CISPO-style).}
To keep the policy from deviating too drastically from a stable, frozen behavior policy $\pi_\beta$, we use a conservative importance-sampling (IS) update. The loss function optimizes only on the terminal action of each trajectory and incorporates a clipped IS ratio $\hat{w}$, an entropy bonus to encourage exploration, and a KL-divergence penalty to regularize the policy update:
\begin{equation}
\mathcal{L} = -\mathbb{E}\big[\hat{w} \cdot \tilde{A} \cdot \log\pi_\theta(a|s)\big] - \eta H(\pi_\theta) + \beta D_{\text{KL}}(\pi_\theta \| \pi_\beta)
\label{eq:loss}
\end{equation}
where $\tilde{A}$ is the standardized advantage. This objective, combined with PEFT techniques (4-bit quantization with LoRA), allows for stable and efficient alignment of the VLM agent on a single 24GB GPU.

\begin{algorithm}[t]
\caption{Policy Alignment via Conservative RL}
\label{alg:rl_alignment}
\begin{algorithmic}[1]
\State \textbf{Input:} Training data $D = \{(x_i, c_i, g_i)\}$, policy $\pi_\theta$, behavior policy $\pi_\beta$
\State Initialize LoRA weights $\theta$; copy to $\beta$

\For{each training step}
    \State Sample minibatch $B \subset D$
    \State Initialize experience buffer $\mathcal{B} \leftarrow []$
    
    \For{each example $(x, c, g)$ in $B$}
        \State // Collect K trajectories and their rewards
        \For{$k = 1, \dots, K$}
            \State // Run loop using fast VLM proxy for KBCS
            \State $p_{\text{final}}, \tau_k \leftarrow$ Run \textbf{Algorithm~\ref{alg:reasoning_loop}} with $\pi_\theta, x, c$
            \State $R_k \leftarrow -(p_{\text{final}} - g)^2$
        \EndFor
        
        \State // Compute advantages using a self-critical baseline
        \State $\bar{R} \leftarrow \frac{1}{K}\sum_k R_k$
        \For{$k = 1, \dots, K$}
            \State $A_k \leftarrow R_k - \bar{R}$
            \State Extract terminal step data $(\log\pi_\theta, \log\pi_\beta, H, \text{KL})$ from $\tau_k$
            \State Append $(A_k, \log\pi_\theta, \log\pi_\beta, H, \text{KL})$ to $\mathcal{B}$
        \EndFor
    \EndFor

    \State // Compute policy gradient loss from buffered experiences
    \State Standardize all advantages $\{A_i\}$ in $\mathcal{B}$ to get $\{\tilde{A}_i\}$
    \State $\mathcal{L}_{\text{PG}} \leftarrow \frac{1}{|\mathcal{B}|} \sum_i \left( -\hat{w}_i \cdot \tilde{A}_i \cdot \log\pi_{\theta,i} \right)$
    \State \quad where $\hat{w}_i = \min(\exp(\log\pi_{\theta,i} - \log\pi_{\beta,i}), c_{\text{clip}})$
    
    \State // Combine losses and update policy
    \State $\mathcal{L}_{\text{reg}} \leftarrow -\eta \cdot \mathbb{E}[H_i] + \beta \cdot \mathbb{E}[\text{KL}_i]$
    \State $\mathcal{L} \leftarrow \mathcal{L}_{\text{PG}} + \mathcal{L}_{\text{reg}}$
    \State Update $\theta$ using gradient descent on $\mathcal{L}$
    \State Periodically, copy weights $\beta \leftarrow \theta$
\EndFor
\end{algorithmic}
\end{algorithm}


\section{Experimental Setup}
\label{sec:setup}

Our experiments are designed to answer three core questions about our framework's ability to produce accurate, faithful, and controllable diagnostic decisions:
\begin{enumerate}[label=(\roman*), topsep=0pt, itemsep=-2pt, partopsep=0pt, parsep=0pt]
    \item \textbf{Does evidence-seeking improve performance?} We compare our interactive agent against a non-interactive VLM baseline to quantify the gains in calibrated accuracy (§\ref{sec:main-results}).
    \item \textbf{Is the reasoning process faithful?} We conduct interventional experiments to verify that the visual evidence identified by the agent is causally linked to its final decision (§\ref{sec:faithfulness_analysis}).
    \item \textbf{Which design choices matter?} We perform a series of ablations to analyze the impact of key components, such as the evidence source, fusion strategy, and RL alignment (§\ref{sec:ablations}).
\end{enumerate}

\paragraph{Datasets and Task.}
Our primary experiments are conducted on a 200-sample subset of the \textbf{VinDr-CXR} dataset~\cite{Nguyen2022VinDrCXR}, covering four common findings: \textit{Pneumothorax}, \textit{Cardiomegaly}, \textit{Pleural effusion}, and \textit{Consolidation}. For each image-finding pair, the agent's task is to predict the probability of the finding's presence. A larger 348-sample set is used for efficiency analysis (§\ref{sec:analysis-efficiency}). We assess out-of-distribution generalization on the \textbf{CheXpert} dataset~\cite{Irvin2019CheXpert} in §\ref{sec:generalization}.

\paragraph{Agent and Baselines.}
Our \textbf{Agent} uses a 4-bit quantized Qwen2.5-VL-3B model as its policy. We evaluate two main versions: an \textbf{Initial Policy} (without RL) and an \textbf{RL-aligned Policy} (trained with our CISPO-style objective). The primary baseline is a \textbf{non-interactive VLM}, which is functionally equivalent to our agent with the \textsc{P\&G} action disabled (\textbf{noP\&G}). This baseline isolates the benefit of the evidence-seeking loop itself. All episodes are capped at $T_{\max}=3$ unless otherwise noted.

\paragraph{Metrics.}
Our primary metrics focus on calibrated accuracy: the \textbf{Brier score} (↓) and \textbf{Expected Calibration Error} (ECE, ↓, 15 bins). To understand agent behavior, we report the rate at which it invokes the \textsc{P\&G} action (\textbf{P\&G Rate}) and the average inference latency (\textbf{WallMS}). Faithfulness is assessed via a causal \textbf{ROI Masking} intervention, where we measure the change in Brier score after occluding the adopted ROI.

\paragraph{Experimental Configurations.}
To thoroughly evaluate our framework, we analyze different configurations by varying two key dimensions: the \textbf{evidence source} used by the \textsc{P\&G} action and the \textbf{fusion strategy} for belief updates.
\begin{itemize}[leftmargin=*, topsep=0pt, itemsep=-2pt]
    \item \textbf{Evidence Source}: We test two sources. (1) \textbf{Prior}: A pre-calibrated evidence score derived from an external model, serving as a reliable but non-visual source. (2) \textbf{KBCS}: The visual evidence score generated in real-time by our KBCS tool, as described in §\ref{sec:method:pg}.
    \item \textbf{Fusion Strategy}: We compare two methods. (1) \textbf{Mix}: A linear interpolation that directly mixes the agent's current belief with the evidence score. (2) \textbf{Gate}: A conservative gating mechanism that only incorporates evidence if it is sufficiently different from the current belief.
\end{itemize}
We refer to configurations by combining these choices, e.g., \textbf{Prior-Mix}. The \textbf{noP\&G} variant serves as our primary baseline.
\section{Main Results: Evidence Improves Calibrated Accuracy}
\label{sec:main-results}

Our central finding is that empowering the VLM agent with the ability to actively seek external evidence leads to significant gains in calibrated accuracy. This answers our first research question (RQ-i). The non-interactive baseline (\textsc{noP\&G}) struggles with this diagnostic task, yielding a high Brier score of 0.491. As shown in Table~\ref{tab:main}, simply enabling the \textsc{P\&G} action allows our agent to substantially improve upon this baseline.

\begin{table}[h!]
\centering
\small 
\setlength{\tabcolsep}{5pt} 
\caption{Main results on the VinDr-200 set. Actively adopting evidence (`Prior-Mix`) dramatically improves performance. The RL-aligned policy learns to use evidence more frequently, achieving the best scores.}
\label{tab:main}
\begin{tabular}{lccccc}
\toprule
Policy (Variant) & Brier ↓ & ECE ↓ & Chk Rate & Adpt Rate & WallMS \\
\midrule
\multicolumn{6}{l}{\textit{Baseline}} \\
ZS / noP\&G     & 0.491 & 0.491 & 0.000    & 0.000    & $\approx$5.3k \\
\midrule
\multicolumn{6}{l}{\textit{Our Agent with Prior-Mix}} \\
Initial Policy   & 0.442 & 0.414 & 0.235 & 0.235 & 13,781 \\
RL-aligned Policy & \textbf{0.403} & \textbf{0.366} & 0.385 & 0.385 & 13,844 \\
\bottomrule
\end{tabular}
\end{table}

The performance lift is driven by two key factors. First, the \textbf{Initial Policy} already learns to leverage evidence when available, reducing the Brier score by 0.049. Second, our \textbf{RL-aligned Policy} learns a more effective evidence-seeking strategy, increasing its \textsc{P\&G} Rate from 23.5\% to 38.5\%. This proactive behavior allows it to correct more initial misjudgments, ultimately achieving a Brier score of \textbf{0.403}—an 18\% relative improvement over the baseline.

\begin{figure*}[t!]
    \centering
    \includegraphics[width=0.7\textwidth]{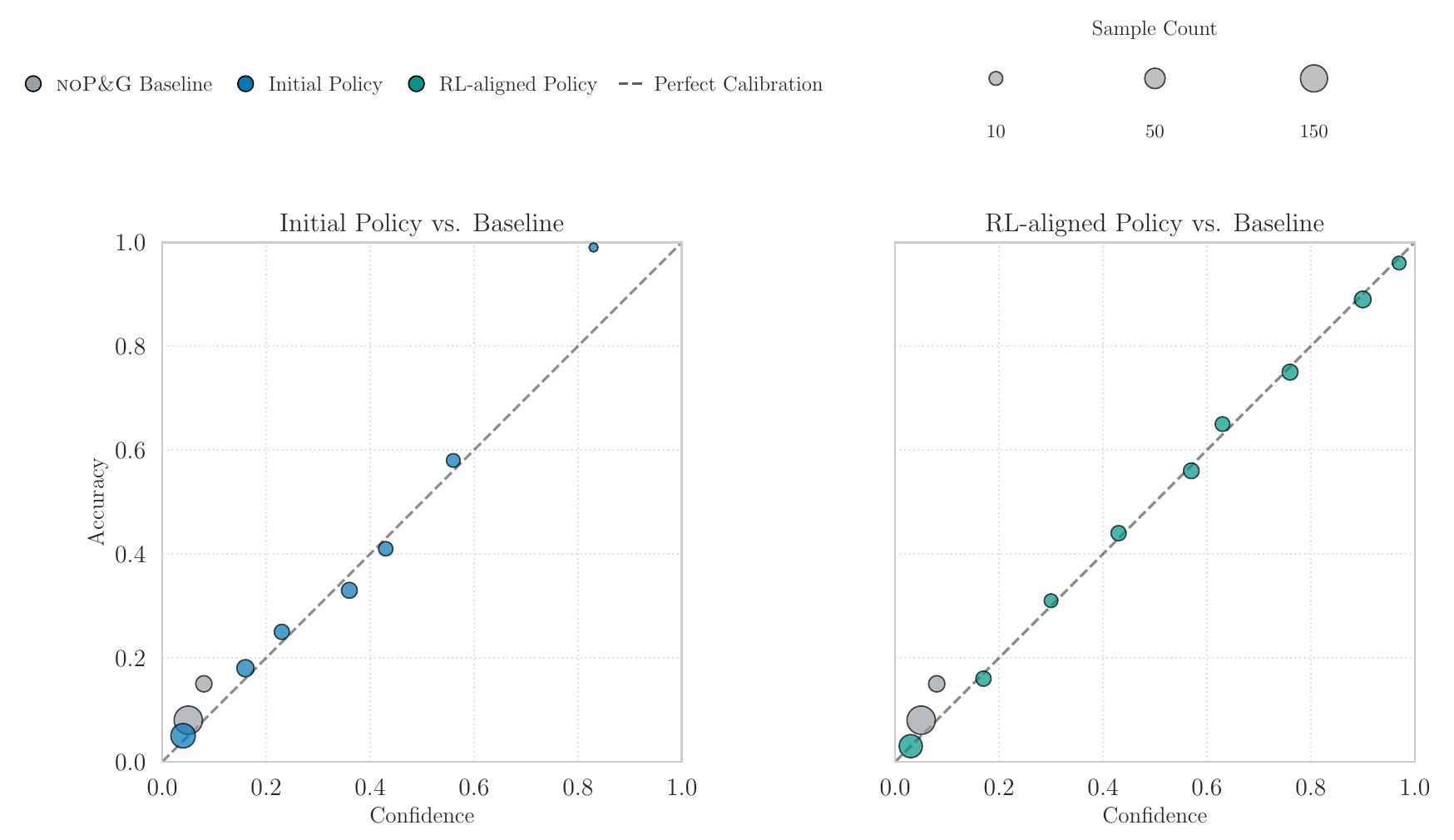}
    \caption{\textbf{Bubble reliability diagrams for the Initial (left) and RL-aligned (right) policies, compared against the \textsc{noP\&G} baseline (gray).} Each bubble's position plots empirical \textbf{Accuracy} against model \textbf{Confidence}, while its size reflects the number of samples in its bin. The baseline's large, low-confidence bubbles show poor calibration. Our agent (blue and green) progressively aligns its bubbles with the ideal diagonal (dashed line) and distributes them more broadly, with the RL-aligned policy (green) achieving the best calibration.}
    \label{fig:reliability}
\end{figure*}

This improvement is fundamentally a story of better calibration, as visualized in the \textbf{bubble reliability diagrams} in Figure~\ref{fig:reliability}. In these diagrams, each bubble represents a confidence bin; its position indicates the average \textbf{Confidence} (x-axis) versus \textbf{Accuracy} (y-axis), while its \textbf{size is proportional to the number of samples} in that bin. A perfectly calibrated model would have its bubbles lying on the dashed diagonal.

The \textsc{noP\&G} baseline (gray bubbles) is severely miscalibrated. Its bubbles lie far from the ideal diagonal and its largest bubbles are anchored in the low-confidence region (0.0-0.2), confirming a model that is both inaccurate and perpetually uncertain. In stark contrast, our agent demonstrates a clear progression towards calibration. The \textbf{Initial Policy} (left plot, blue) shifts its bubbles closer to the diagonal and begins to populate the mid-confidence range. This transformation is perfected by the \textbf{RL-aligned Policy} (right plot, green). Its bubbles align tightly with the ideal diagonal, and their sizes indicate a healthy distribution of predictions across the full confidence spectrum. This visual journey from a timid, miscalibrated model to a confident, reliable one directly explains the substantial ECE reduction from 0.491 to 0.366 noted in Table~\ref{tab:main}.

\section{Ablation Studies and Component Analysis}
\label{sec:ablations}

We now dissect our framework's performance by analyzing its core components to answer our third research question (RQ-iii). We investigate how the evidence source and fusion strategy impact performance (§\ref{sec:analysis-ablation}), the trade-off between performance and efficiency (§\ref{sec:analysis-efficiency}), and the controllability of the agent's belief update mechanism (§\ref{sec:analysis-gating}).

\subsection{Evidence Source and Fusion Strategy}
\label{sec:analysis-ablation}

\paragraph{Well-calibrated evidence is essential.}
To isolate the impact of evidence quality and the belief fusion logic, we conducted a comprehensive ablation study. The results, presented in Table~\ref{tab:abl-source}, underscore a critical finding: performance gains are contingent not just on seeking evidence, but on seeking \textit{high-quality, calibrated} evidence and integrating it appropriately.

The `Prior-Mix` variant, which updates the agent's belief with a calibrated score, consistently delivers the best performance. For the \textbf{Initial Policy}, it reduces the Brier score from 0.491 to \textbf{0.442}. After RL alignment, this gain is even more pronounced, with the score dropping to \textbf{0.403}.

Conversely, using an uncalibrated source can be actively harmful. The `KBCS-Mix` variant, when used by the \textbf{RL-aligned Policy}, demonstrates clear negative transfer, increasing the Brier score to 0.499 compared to the 0.491 of the \textsc{noP\&G} baseline. Furthermore, the conservative `Gate` strategy proves overly cautious; its strict update rule prevents the agent from ever adopting evidence, making it functionally equivalent to the \textsc{noP\&G} baseline.

\begin{table}[htbp]
\centering
\small
\caption{Ablation on evidence source and fusion strategy. The calibrated `Prior-Mix` variant consistently provides the best performance. Naively mixing uncalibrated KBCS evidence leads to negative transfer.}
\label{tab:abl-source}
\begin{tabular}{llccc}
\toprule
Policy & Variant & Brier Score (↓) & ECE (↓) & Adopt Rate \\
\midrule
\multirow{4}{*}{Initial} 
 & \textsc{noP\&G} & 0.491 & 0.491 & 0.000 \\
 & KBCS-Gate   & 0.480 & 0.467 & 0.000 \\
 & KBCS-Mix    & 0.483 & 0.470 & 0.235 \\
 & \textbf{Prior-Mix}   & \textbf{0.442} & \textbf{0.414} & 0.235 \\
\midrule
\multirow{3}{*}{RL-aligned} 
 & \textsc{noP\&G} & 0.491 & 0.491 & 0.000 \\
 & KBCS-Mix    & 0.499 & 0.498 & 0.350 \\
 & \textbf{Prior-Mix}   & \textbf{0.403} & \textbf{0.366} & 0.385 \\
\bottomrule
\end{tabular}
\end{table}

\subsection{Efficiency of the RL-aligned Policy}
\label{sec:analysis-efficiency}

We also analyzed the agent's efficiency by varying the maximum allowed interaction steps, $T_{\max}$, for the RL-aligned policy. The results, summarized in Table~\ref{tab:steps}, reveal two key findings. First, performance is not monotonic with the step budget; the best Brier score (\textbf{0.383}) is achieved at $T_{\max}=4$. Second, and more importantly, the agent remains highly efficient even with a larger budget. For instance, at $T_{\max}=4$, the average number of steps taken is only 1.34. This demonstrates that the policy has learned to terminate early when confident and use additional steps judiciously only when necessary, avoiding wasteful interactions.

\begin{table}[h!]
\centering
\small
\caption{Performance of the RL-aligned policy versus the maximum step budget ($T_{\max}$). The agent achieves the best Brier score at $T_{\max}=4$ while maintaining a low average step count, demonstrating learned efficiency.}
\label{tab:steps}
\begin{tabular}{cccc}
\toprule
Max Steps ($T_{\max}$) & Brier Score ↓ & ECE ↓ & Avg. Steps \\
\midrule
1 & 0.392 & 0.335 & 1.00 \\
2 & 0.391 & 0.332 & 1.26 \\
3 & 0.400 & 0.363 & 1.23 \\
4 & \textbf{0.383} & 0.337 & 1.34 \\
\bottomrule
\end{tabular}
\end{table}

\subsection{Controllability of the Gating Mechanism}
\label{sec:analysis-gating}
\paragraph{Evidence quality trumps adoption quantity.}
Finally, we investigated whether the belief update process is controllable and if simply adopting more evidence improves performance. We swept the threshold $\tau$ of the `Gate` fusion strategy, which governs whether new evidence is integrated.

\begin{figure}[!ht]
    \centering
    \includegraphics[width=0.7\columnwidth]{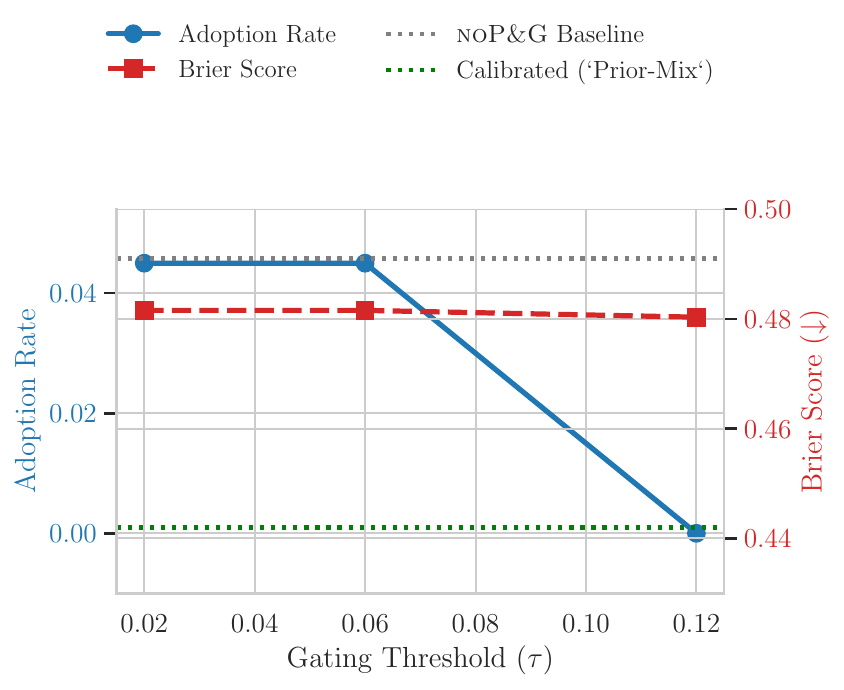}
    \caption{Effect of gating threshold $\tau$ on adoption rate and performance for the uncalibrated KBCS-Gate variant. While the adoption rate is controllable, performance remains stagnant, highlighting the importance of evidence quality over quantity.}
    \label{fig:sweep-gate}
\end{figure}

The results, summarized in Figure~\ref{fig:sweep-gate}, show that the adoption mechanism is indeed controllable. A less strict threshold ($\tau=0.02$) allows the agent to update its belief in 4.5\% of cases. As we increase the threshold, the adoption rate drops to zero. However, this modulation did not translate into meaningful performance gains. The Brier score remained stagnant around 0.480, failing to approach the performance of the calibrated `Prior-Mix` variant. This confirms a key insight: for an uncalibrated evidence source, merely controlling the \textit{quantity} of adopted evidence is insufficient. The \textit{quality} of the evidence, achieved through calibration, is the dominant factor for performance improvement.

\section{Faithfulness and Generalization Analysis}
\label{sec:faithfulness_generalization}

Having established our agent's core performance and analyzed its components, we now address two advanced properties: the faithfulness of its reasoning process (RQ-ii) and its ability to generalize to new data distributions.

\subsection{Faithfulness Analysis: From Correlation to Causation}
\label{sec:faithfulness_analysis}

A faithful process requires that the evidence cited genuinely contributes to the final decision. We investigate this by first exposing the limitations of standard correlation-based methods, which then motivates our more rigorous agent-level causal intervention.

\paragraph{The Limits of Correlation: Occlusion-Drop on an Adaptive Tool.}
A common method to assess faithfulness is Occlusion-Drop analysis, measuring the score drop when a relevant region is masked. We applied this directly to our KBCS tool. As shown in Table~\ref{tab:occlusion}, masking human-annotated Ground-Truth (GT) ROIs causes a score drop, suggesting GT regions contain critical information.

However, a stark contradiction appears when masking the ROI predicted by the KBCS tool itself: the occlusion drop is \textbf{zero}. This paradoxical result highlights a weakness of this method on adaptive tools. Our KBCS is designed to re-localize evidence; masking its first-choice ROI simply causes it to select the next-highest peak on its subsequent run, yielding a nearly identical score. This renders Occlusion-Drop ineffective for assessing the tool in isolation.

\begin{table}[h!]
\centering
\small
\caption{Occlusion-Drop analysis on the KBCS tool. Masking the tool's own Predicted ROI yields a zero drop, demonstrating the method's failure due to the tool's adaptive re-localization.}
\label{tab:occlusion}
\begin{tabular}{lcccc}
\toprule
ROI Source & Real Drop (↓) & Rand Drop (↓) & Diff (↑) & Cohen's d (↑) \\
\midrule
GT ROI     & 0.133 & 0.084 & 0.049 & 0.031 \\
Pred. ROI  & \textbf{0.000} & \textbf{0.000} & \textbf{0.000} & \textbf{0.000} \\
\bottomrule
\end{tabular}
\end{table}

\paragraph{From Correlation to Causation: Agent-Level Intervention.}
The right question is not "what can the tool find?", but rather "what evidence does the agent \textit{actually use}, and does that use causally affect its final decision?". Our framework allows for a true causal intervention. We identified a cohort of \textbf{N=77} cases where our best model (RL-aligned with Prior-Mix) actively adopted evidence from the KBCS. It is exclusively on this "adopted" cohort that we perform our intervention: we mask the exact ROI the agent used and re-evaluate its performance.

The results in Table~\ref{tab:interventional} are unequivocal. After masking the specific visual evidence the agent chose to act upon, its Brier score significantly increased ($\Delta=\text{+0.029}$), indicating a substantial performance degradation. This demonstrates that the adopted ROI is not a post-hoc rationalization but is causally integral to the agent's final decision, validating the faithfulness of our framework.

\begin{table}[h!]
\centering
\caption{Causal faithfulness test. On the N=77 subset of cases where evidence was adopted, masking the identified ROI significantly degrades the agent's performance, confirming a causal link.}
\label{tab:interventional}
\resizebox{0.5\columnwidth}{!}{%
\begin{tabular}{lcc}
\toprule
Intervention Setting & Brier Score (↓) & ECE (↓) \\
\midrule
Before (Original Image) & 0.441 & 0.414 \\
After (Masked ROI)      & 0.470 \small{(+0.029)} & 0.452 \small{(+0.038)} \\
\bottomrule
\end{tabular}%
}
\end{table}

\subsection{Generalization Under Distribution Shift}
\label{sec:generalization}

To assess our framework's adaptability, we evaluated the Initial Policy's performance on the \textbf{CheXpert} dataset. As shown in Table~\ref{tab:interventional}, the agent's core evidence-seeking behavior remains effective.

More importantly, we tested if performance could be enhanced with minimal, test-time adaptation. By applying a simple, per-concept temperature scaling factor ($T=4.0$), fitted on a small target-domain calibration set, we observed a consistent improvement in calibrated accuracy (Brier $\Delta=-0.006$, ECE $\Delta=-0.009$). This demonstrates that our agent's reasoning process can be effectively re-calibrated for a new domain \textit{without any model retraining}, highlighting the modularity and practicality of our framework.
\begin{table}[h!]
\centering
\caption{Generalization to CheXpert ($N=600$). A simple temperature-scaling overlay at test time improves calibration on the new domain without any model retraining.}
\label{tab:chexpert}{%
\begin{tabular}{lcccc}
\toprule
Setting & Brier ↓ & ECE ↓ & P\&G Rate & WallMS (k) \\
\midrule
Baseline (Source Calibration) & 0.271 & 0.069 & 0.273 & 27.3 \\
\textbf{+ Target-domain Calib.} & \textbf{0.265} & \textbf{0.060} & 0.273 & 24.0 \\
\bottomrule
\end{tabular}%
}
\end{table}

\section{Conclusion}
\label{sec:conclusion}

We introduced an interactive agent that externalizes its reasoning into an auditable sequence of actions. By learning to strategically seek and ground its beliefs in visual evidence, our agent achieves significant gains in calibrated accuracy (18\% Brier score reduction) over a non-interactive baseline. More importantly, we validated its faithfulness with a rigorous causal intervention: masking the agent's chosen evidence systematically degrades performance ($\Delta\text{Brier}=+0.029$), proving its reasoning is not a post-hoc fabrication.

This work champions an action-centric view of explainability. Instead of interpreting a model's internal states, we optimize its external behavior, forcing the agent to \textit{justify} its conclusions through interaction. This shift from passive interpretation to active, verifiable reasoning offers a practical blueprint for building AI systems that are not only accurate but also demonstrably trustworthy.

\bibliographystyle{plainnat}
\bibliography{main}

\end{document}